# Intelligent road crack detection and analysis based on improved YOLOv8

Haomin Zuo
School of Electronics and Communication Engineering
Sun Yat-sen University
Guangzhou, China
2519816821@qq.com

Zhengyang Li*
Master of Science in Computer Science
DigiPen Institute of Technology
Redmond, WA, USA
* Corresponding author: levey.lee@gmail.com

Jiangchuan Gong
Hebei Normal University
Shijiazhuang, Hebei, China
jackgong151823@gmail.com

Zhen Tian
James Watt School of Engineering
University of Glasgow
Glasgow, UK
2620920Z@student.gla.ac.uk

*Abstract*—As urbanization speeds up and traffic flow increases, the issue of pavement distress is becoming increasingly pronounced, posing a severe threat to road safety and service life. Traditional methods of pothole detection rely on manual inspection, which is not only inefficient but also costly. This paper proposes an intelligent road crack detection and analysis system, based on the enhanced YOLOv8 deep learning framework. A target segmentation model has been developed through the training of 4029 images, capable of efficiently and accurately recognizing and segmenting crack regions in roads. The model also analyzes the segmented regions to precisely calculate the maximum and minimum widths of cracks and their exact locations. Experimental results indicate that the incorporation of ECA and CBAM attention mechanisms substantially enhances the model's detection accuracy and efficiency, offering a novel solution for road maintenance and safety monitoring.

## I. Introduction

As urbanization accelerates and traffic flow continues to increase, the issues of pavement potholes and other road surface diseases are becoming increasingly prominent, posing a serious threat to traffic safety and the longevity of road services. The conventional method for detecting pavement potholes primarily depends on manual inspection, which is not only inefficient and costly but also susceptible to omissions and misidentifications. In recent years, with the swift advancement of computer vision and deep learning technologies, image-based target detection algorithms have progressively emerged as the leading approach for identifying pavement potholes.

This study[1] presents a Res50-SimAM-ASPP-Unet model for high-resolution remote sensing image segmentation, integrating ResNet50, SimAM attention, and ASPP to improve feature extraction and context understanding. Results on LandCover.ai demonstrate high performance, with a Mean Intersection over Union (MIOU) of 81.1%, accuracy of 95.1%, and an F1 score of 90.45%. Another paper[2] introduces a visual state-space model that utilizes wavelet guidance, an enhanced U-structure, and patch resampling for improved skin lesion segmentation. Lin et al. [3-4] propose a lightweight visual SLAM framework designed for dynamic object filtering and real-time obstacle avoidance in autonomous vehicles, ensuring safe navigation. Furthermore, SLAM2 integrates geometry, semantics, and dynamic object maps for indoor environments, employing deep learning for real-time multi-mode modeling, thereby enhancing dynamic obstacle tracking and scene understanding.

The Lu research team presented two innovative studies[5-6]. The first, CausalSR, combined structural causal models with counterfactual inference to enhance super-resolution reconstruction, reducing artifacts and distortions in complex scenarios like medical and satellite imaging. The second study developed a framework for automated pavement texture extraction and evaluation using computer vision and machine learning, which quantifies road quality metrics to aid in road maintenance and infrastructure monitoring. Deep Learning Technology has achieved a breakthrough in traditional detection methods in road engineering. The multi-view stereo reconstruction and lightweight deep learning framework proposed by Dan et al. [7], as well as the combination of U-Net segmentation with interactive image processing in the[8]study, have both improved the efficiency and accuracy of pavement evaluation, and have improved the accuracy of pavement evaluation, promote the digitalization and intellectualization of road engineering.

In order to solve these problems, researchers have extensively improved the YOLO series of algorithms. For example, literature [9] proposed a pavement pothole detection algorithm based on improved YOLOv5s, which significantly improved the model's ability to detect small targets and feature extraction accuracy by introducing GFPN module and CA module. Literature [10], on the other hand, further improves the detection accuracy and checking rate of YOLOv5 by introducing BiFPN attention mechanism and DIoU loss function. Literature [11] proposed a pavement pothole detection algorithm based on the improved YOLOv8, which significantly improves the detection accuracy and lightness of the model by introducing the CPCA attention mechanism and MPDIoU loss function.

While existing studies have enhanced the performance of the YOLO series algorithms to some extent, the current models still require further optimization, particularly in terms of complex background, small target detection, and real-time performance. Consequently, this paper proposes an improved YOLOv8 model, designed to further improve the accuracy and efficiency of pavement pothole detection. This is achieved by optimizing feature extraction, incorporating an efficient attention mechanism, and refining the loss function.

## II. METHODOLOGIES

### A. YOLOv8 Network Architecture

YOLOv8 is a state-of-the-art target detection technology that inherits the success of previous YOLO versions in target detection tasks and realizes significant performance and flexibility improvements, combining superior accuracy and speed. Compared to its predecessor, YOLOv8 introduces a number of innovations, including a new backbone network, an Anchor-Free detection header, and an improved loss function, making it efficient on a wide range of hardware platforms, from CPUs to GPUs, making it ideal for a wide range of object detection tasks.

The network structure of YOLOv8 consists of three parts: Backbone, Neck and Head. Backbone adopts the improved CSPDarknet structure, which enhances the feature extraction capability through C2f module and introduces SPPF module to improve the inference speed. The Head part uses the Anchor-Free mechanism to directly predict the centroid and width of the target, and uses the CIoU loss function to optimize the bounding box regression accuracy, which is responsible for the final target detection and classification task.The structure of the YOLOv8 network is as follows: its specific structure is shown in Figure 1 below.

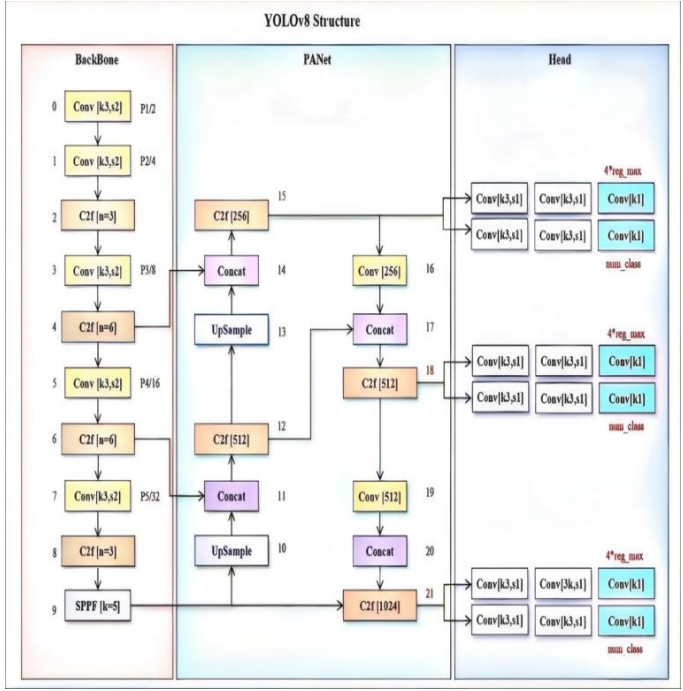

Figure 1.  YOLOv8 network structure

### B. Attention mechanism

1.ECA Attention Mechanism

ECA (Efficient Channel Attention) attention mechanism can significantly enhance the model's response to important features by dynamically adjusting the key channel weights in the convolutional attention features. Its core working principle includes: firstly, global average pooling operation is performed on the feature map to obtain global context information; then channel weights are generated through 1D convolution operation; finally, these weights are applied to the original feature map to generate the weighted output feature map.The advantage of the ECA mechanism lies in its high efficiency and lightweight characteristics, which avoids complex global attention computation, and at the same time, it is able to adaptively adjust the channel weights. weights. This makes ECA perform well in lightweight models and significantly improves the performance of the model. Its specific structure is shown in Figure 2.

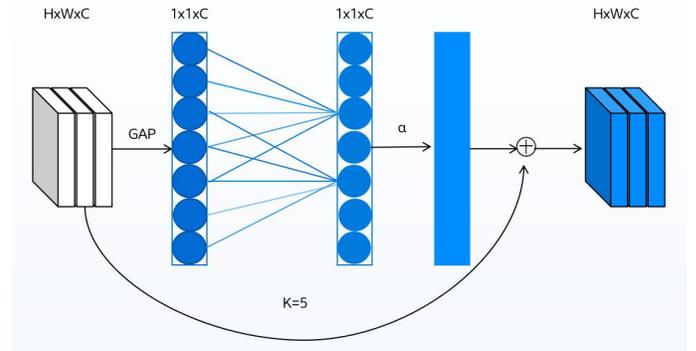

Figure 2.  ECA module

The ECA attention mechanism enhances the representation of the feature map through three steps. First, the global information of each channel is extracted by global average pooling to prepare for the computation of channel weights. Then, a 1-dimensional convolution of size k and a Sigmoid activation function are utilized to generate the channel weights w, which enables local cross-channel interactions and captures inter-channel dependencies. Finally, the obtained weights w are multiplied element-by-element with the original feature map to generate the final output feature map.

2.CBAM Attention Mechanism

CBAM (Convolutional Block Attention Module) is a composite attention mechanism that combines channel attention and spatial attention, aiming to enhance the feature representation capability of convolutional neural networks. The channel attention module obtains the global statistical information of each channel through global average pooling and global maximum pooling operations, and processes this information using two fully connected layers to generate channel weights and strengthen the influence of important channels. The spatial attention module, on the other hand, based on the output of channel attention, further learns the importance of spatial locations by performing global pooling on the feature map and generates a spatial weight map to highlight the features of important spatial regions. The combination of these two can effectively improve the model's ability to recognize the importance of features. Its specific working principle is shown in Figure 3 below.

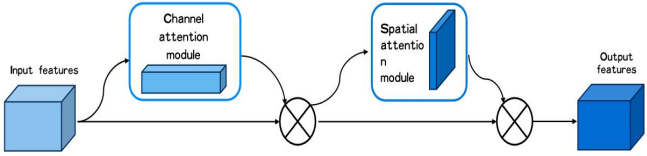

Figure 3. CBAM module

The Channel Attention Module (CAM) aims to enhance the representation of each channel of the feature map, and its main steps include: firstly, obtaining the maximum and average feature values of each channel through global maximum pooling and global average pooling to capture the global information; then using the shared fully connected layer to learn the attention weights of each channel, to further abstract and encode the features; then applying the Sigmoid activation function is applied to restrict the weights between 0 and 1 to reflect the importance of the channels; finally, the computed attentional weights are multiplied element-by-element with the original feature maps, such that the features of the important channels are augmented while the features of the unimportant channels are suppressed.

The Spatial Attention Module (SAM) emphasizes the importance of different locations in the image through a series of steps. First, a global pooling operation is performed to obtain the maximum and average feature maps; these two feature maps are then spliced in the channel dimension to form a richer feature map. Next, the spliced feature maps are downscaled using a 7 × 7 convolutional layer to learn the dependencies between spatial locations. Subsequently, a Sigmoid activation function is applied to generate spatial attention weights reflecting the importance of each location. Finally, feature weighting is achieved by element-by-element multiplication with the original feature map, which allows features at important locations to be enhanced, while features at unimportant locations are suppressed.

## III. EXPERIMENTAL RESULTS

### A. Experimental design

This paper describes the dataset preparation and training process of the road crack detection program. 4029 road crack related images were collected through the network, and each image was labeled with segmentation results and categories using the Labelimg tool. Finally, the dataset is divided into training set (3717 images), validation set (200 images) and test set (112 images).

The experimental environment of this paper is based on Window11 operating system, the deep learning framework is PyTorch, the CPU used is 13th Gen Intel(R) Core(TM) i9-13900HX, 2.20 GHz, and the GPU selected is NVIDIA's RTX 4060 with 16G video memory.

In order to provide a comprehensive and accurate comparison with current state-of-the-art methods, this experiment employs several key evaluation metrics to validate the performance of road crack segmentation. These evaluation metrics include Recall (R), Precision (P), and Accuracy (A). The formulas are shown below.

$$Recall = \frac{TP}{TP+FN} \quad (1)$$

$$P = \frac{TP}{TP+FP} \quad (2)$$

$$A = \frac{TP+TN}{TP+TN+FP+FN} \quad (3)$$

### B. Analysis of results

In this paper, we conduct a thorough analysis of the specific enhancement effect that the attention mechanism module has on the performance of the YOLOv8 model, using well-designed ablation experiments. During these experiments, we sequentially integrate two advanced attention mechanism modules, ECA (Efficient Channel Attention) and CBAM (Convolutional Block Attention Module), into the YOLOv8 model. We then comparatively analyze the performance of the model under different configurations. This comparative analysis aims to clearly demonstrate the independent contribution of each module to the improvement in model performance. The experimental results indicate that both the ECA module and the CBAM module effectively enhance the segmentation accuracy and feature extraction capability of the model, thereby strongly validating the effectiveness and practicality of the proposed improvement strategy.

TABLE I. COMPARISON OF EXPERIMENTAL RESULTS

| YOLOv8 | ECA | CBAM | Recall | P | A |
|---|---|---|---|---|---|
| ✓ | x | x | 78.45 | 80.89 | 79.85 |
| ✓ | ✓ | x | 82.86 | 82.34 | 81.86 |
| ✓ | x | ✓ | 84.47 | 85.33 | 84.34 |
| ✓ | ✓ | ✓ | **89.47** | **92.25** | **91.34** |

Referring to the experimental data in table I, the ECA and CBAM attention mechanisms significantly enhance the performance of the YOLOv8 model. When YOLOv8 is utilized in isolation, Recall, Precision, and Accuracy are 78.45%, 80.89%, and 79.85%, respectively. With the integration of the ECA module, Recall and Accuracy improve to 82.86% and 81.86%, respectively; and upon further integration of the CBAM module, Recall, Precision, and Accuracy are further enhanced to 84.47%, 85.33%, and 84.34%. The optimal combination, which involves the simultaneous integration of ECA and CBAM, results in model performance metrics of 89.47%, 92.25%, and 91.34%. This demonstrates that the combination of the two mechanisms enhances feature representation from both channel and spatial dimensions, effectively improves detection accuracy without substantially increasing computational complexity, and is well-suited for real-time application scenarios.

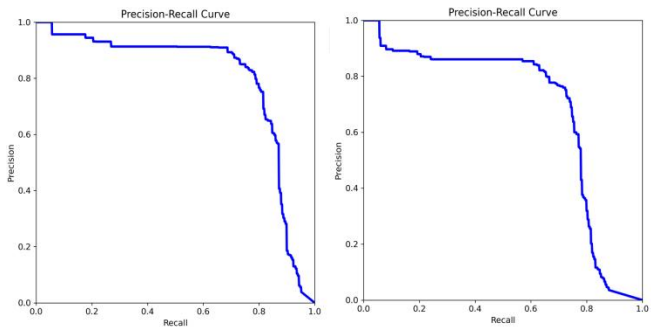

Figure 4. PR curve

Figure 4 illustrates the Precision-Recall (PR) curve for the classification model, which is a key indicator of model performance. The horizontal axis represents the recall rate, while the vertical axis corresponds to the precision rate. Ideally, the curve should approach the upper left corner to signify high precision and recall. The curve indicates that as the recall rate increases, the precision rate decreases; at lower recall rates, the precision rate is nearly 1, suggesting that the model predominantly identifies positive samples among the detected ones. Observing the curve results from the figure above: the average precision for localization is 0.799, and for segmentation, it is 0.685. These results remain quite satisfactory.

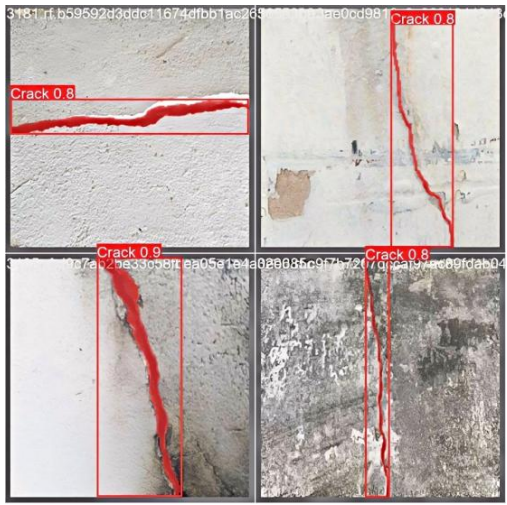

Figure 5. Comparison of Segmentation Effect Graphs

Figure 5 illustrates the outcomes of applying the target detection model to identify wall cracks. In each subfigure, the red line highlights the location of the crack detected by the model, and the number adjacent to it signifies the model's confidence score for that particular crack. As depicted in the figure, there are variations in the model's effectiveness at detecting cracks across different backgrounds. In more uniform and clean environments, such as the upper left and upper right figures, the model accurately identifies cracks with confidence scores of 0.8 and 0.9, respectively, suggesting that the model is more dependable in recognizing these cracks. Conversely, in scenarios where the background is intricate or the crack features are subtle, as seen in the lower right figure, the model's confidence score is also 0.8; however, the detection lines do not align as closely with the actual cracks as in the other subfigures. This discrepancy may suggest that the model's ability to recognize cracks in such complex backgrounds is limited. Overall, these results affirm the model's efficacy in crack detection under various conditions and underscore the necessity for enhanced detection precision in complex backgrounds. This data is crucial for further refining the model and bolstering its resilience in real-world applications.

## IV. CONCLUSION

This paper presents YOLOv8 with enhanced performance in detecting pavement potholes, thanks to ECA and CBAM attention mechanisms. The model's Recall, Precision (P), and Accuracy (A) significantly improve, especially when both mechanisms are combined, reaching 89.47%, 92.25%, and 91.34% respectively. Ablation experiments confirm the individual contributions of ECA and CBAM to performance gains, demonstrating their effectiveness. The results indicate that introducing advanced attention mechanisms can greatly enhance detection accuracy and robustness without slowing down the model, offering valuable insights for real-world applications. Future research could further investigate the combination of various attention mechanisms with YOLOv8 for improved detection outcomes.